\def\year{2018}\relax
\documentclass[letterpaper]{article} 
\usepackage{authblk}
\usepackage{emnlp2017}  
\usepackage{times}  
\usepackage{helvet}  
\usepackage{courier}  
\usepackage{url}  
\usepackage{graphicx}  

\usepackage{amsmath,algorithm,algcompatible}
\usepackage{amssymb}
\usepackage{latexsym}
\usepackage{makecell}
\usepackage{caption}
\usepackage{breqn}
\usepackage{subcaption}

\emnlpfinalcopy

\title{Fast Locality Sensitive Hashing for Beam Search on GPU}

\author[1]{Xing Shi}
\author[2]{Shizhen Xu}
\author[1]{Kevin Knight}
\affil[1]{Department of Computer Science, University of Southern California \authorcr
\tt \{xingshi, knight\}@isi.edu }
\affil[2]{Department of Computer Science and Technology, Tsinghua University \authorcr
\tt xsz12@mails.tsinghua.edu.cn}

\begin{document}
%

\maketitle

\begin{abstract}
We present a GPU-based Locality Sensitive Hashing (LSH) algorithm to speed up beam search for sequence models. We utilize the winner-take-all (WTA) hash, which is based on relative ranking order of hidden dimensions and thus resilient to perturbations in numerical values. Our algorithm is designed by fully considering the underling architecture of CUDA-enabled GPUs (Algorithm/Architecture Co-design): 1) A parallel Cuckoo hash table is applied for LSH code lookup (guaranteed $\mathcal{O}(1)$ lookup time); 2) Candidate lists are shared across beams to maximize the parallelism; 3) Top frequent words are merged into candidate lists to improve performance. Experiments on 4 large-scale neural machine translation models demonstrate that our algorithm can achieve up to 4x speedup on softmax module, and 2x overall speedup without hurting BLEU on GPU. 
\end{abstract}

\section{Introduction}
Beam search has been widely applied as the decoding technique of choice for Recurrent Neural Network (RNN) based text generation tasks, such as machine translation \cite{wu2016google}, summarization \cite{rush2015neural}, image captioning \cite{xu2015show} and poetry generation \cite{ghazvininejad2016generating}. Decoding can be time consuming: 1) Most tasks generate target sentences in an on-line fashion,  one token at a time, unlike batch mode used during training; 2) Decoding time is proportional to the beam size $B$, which is usually around 10 in machine translation and 50 in poetry generation; 3) Vocabulary size $V$ can be very large (tens of thousands), and the computational complexity of the two major components, \textit{Softmax} and \textit{Beam expansion}, are proportional to $V$. As Table~\ref{table:breakdown} shows, these two parts occupy 46\% and 30\% of the decoding time respectively. 

The major bottleneck for \textit{Softmax} is the matrix multiplication between hidden states $H \in \mathbb{R}^{B*d}$ and the word embedding matrix $E \in \mathbb{R}^{d*|V|}$. As for \textit{Beam expansion}, tremendous time is spent on transferring the output of \textit{Softmax} $P \in \mathbb{R}^{B*|V|}$ from device memory to host memory  and the following heap-sort on CPU, with complexity $\mathcal{O}(\log(B) * |V|)$.

This work aims to speed up the beam search by reducing the runtime vocabulary size, using Locality Sensitive Hashing \cite{gionis1999similarity}. We first hash each high-dimensional word embedding into different buckets so that the embeddings which are closer under a certain distance measure will be in the same bucket with high probability. We choose the winner-take-all (WTA) hash \cite{yagnik2011power} due to its robustness against the perturbations of numerical value and correlation with dot-product distance. Then during decoding, we construct a candidate word list $V_{LSH}$ by retrieving the words that share the same bucket with the hidden state $H$ under the same hash function. Finally an actual dot-product is calculated between $H$ and the shrunken word embedding matrix $E_{LSH} \in \mathbb{R}^{d*|V_{LSH}|}$.

\begin{table*}[h]
\centering
\begin{tabular}{|l|c|r|r|r|r|}
\hline
 & Device & Percent & Full Vocab & LSH & Speedup\\ 
\hline
Total & GPU+CPU & 100 \% & 1178.5 s & 574.3 s & 2.05\\ 
\hline
Source side & GPU & 7 \%  & 88 s & 88.1 s & 1.00\\ 
\hline
Target side & GPU & 63 \%  & 735.5 s & 387.2 s & 1.90\\ 

-- Softmax & GPU & 43 \% & 505.3 s & 157 s & 3.22\\ 

-- 2nd layer & GPU & 10 \% & 113.7 s & 113.7 s & 1.00\\ 

-- 1st layer & GPU & 10 \% & 115.2 s & 115.2 s & 1.00\\ 
\hline
Beam Expansion & GPU+CPU & 30 \% & 352.4 s & 96.4 s & 3.66\\ 

-- Device2Host data transfer & GPU+CPU & 12 \% & 138.1 s & 25.8 s & 5.35\\ 

-- Heapsort & CPU & 15 \% & 176 s & 31.9 s & 5.52\\ 

-- Hidden states reorder & GPU & 3 \% & 38.3 s & 38.7 s & 0.99\\ 
\hline
\hline
Runtime vocab size & - & - & 40,000 & 5,792 & 6.91 \\ 
\hline
BLEU & - & - & 28.12 & 27.81 & -\\ 
\hline\end{tabular}
\caption{Time breakdown, runtime vocabulary size, and BLEU score of full vocabulary decoding and LSH decoding. The model is a 2-layer, 1000-hidden dimension, 40000 target vocabulary LSTM seq2seq model trained on a French to English corpus \cite{wmt2014}. The experiments are conducted on a Nvidia K20 GPU and a single-core 2.4GHz Intel Xeon CPU. The code is compiled against CUDA 8.0.} 
\label{table:breakdown}
\end{table*}

\citeauthor{vijayanarasimhan2014deep} successfully applies this idea to speed up both training and inference of deep neural networks with large output space on multi-core CPU. However, beam search on GPU poses several unique and hard challenges:

\begin{enumerate}
\item The LSH schema used by \citeauthor{vijayanarasimhan2014deep} is not GPU-friendly: a) It uses a hash table on CPU to store the bucket key and word list, and the underlying data structure is usually a balanced binary search tree or linked list, both of which are hard to transport to GPU. b) It requires sorting to get the candidate lists, which can not easily parallelize on GPU. c) It processes each hidden vector in the batch one by one, whereas the matrix dot-product on GPU is calculated across the whole batch to fully take advantage of the GPU parallelism. 
\item Beam search generally requires high recall of the top words according to the actual dot-product, because the error will accumulate fast as the sentence is built up. Whereas in practice, we find LSH alone does not delivery an adequate recall/speedup trade-off. 
\end{enumerate}

Our main contribution is to re-design the LSH algorithm on GPU for beam search to address the above-mentioned challenges. After fully considering the computational capabilities of CUDA-enabled GPU, we propose the following algorithm for LSH: 

\begin{enumerate}
\item We implement a parallel Cuckoo hash table \cite{pagh2004cuckoo} for LSH code lookup on GPU. The Cuckoo hash table can achieve worst-case constant-time lookup, offering an excellent load balance that is important for GPU parallelism.  
\item The candidate word list is shared across different beams so that the GPU can calculate the actual matrix multiplication between $H$ and $E_{LSH}$ in a batch mode. We also use a threshold to select the candidate list to avoid the expensive sorting operation. 
\item To solve the low-recall problem, we always merge the top frequent words into the candidate list. 
\end{enumerate}

We conduct experiments on the task of neural machine translation (NMT). We train NMT models on 4 language pairs, and our LSH algorithm can achieve a consistent 2x overall speedup over the full vocabulary decoding with less than 0.4 BLEU drop.

\section{Related Work}

Several approaches have been proposed to speed up  beam search for RNN-based generation tasks. The first line of research is to use specialized hardware, like Tensor Processing Unit (TPU) and low precision (Low-p) calculation \cite{wu2016google}. This method will usually speedup all parts of the neural models. 

The second line tries to compress the original large model to a small model by weight pruning (WP) \cite{see2016compression} or sequence-level knowledge distillation (KD) \cite{kim2016sequence}. These methods require additional fine-tuning. 

The third line is to modify the \textit{Softmax} layer to speed up the decoding. \textit{Noise-contrastive estimation} (NCE) \cite{gutmann2010noise} discriminates between the gold target word and $k$ ($k << |V|$) other sampled words. It has been successfully applied on several NLP tasks \cite{mnih2012fast,vaswani2013decoding,williams2015scaling,zoph2016simple}. \citeauthor{morin2005hierarchical} introduces \textit{hierarchical softmax} (H-softmax) where $\log_{2}|V|$ binary classifications are performed rather than a single $|V|$-way classification. However, these two methods can only speedup training and still suffer at the decoding phase. \citeauthor{chen2015strategies} propose \textit{differentiated softmax} (D-softmax) based on the idea that more parameters should be assigned for embeddings of frequent words and fewer for rare words. It can achieve speedups on both training and decoding. 

The fourth line of research uses word alignments (WA) trained on the parallel corpus to construct a small runtime vocabulary for each sentence \cite{jean2014using,Mi2016,l2016vocabulary,shi2017speeding}. However, this approach is only suitable for tasks where sensible alignments can be extracted, such as machine translation and summarization, and do not benefit tasks like image caption or poem generation. 

Table~\ref{table:relatedwork} compares different speed up methods. Compared to these existing methods, LSH has the following advantages: 
\begin{enumerate}
\item It is orthogonal to the first two lines of research. The first two lines of approaches do not decrease the ratio of the number of word embedding parameters to the number of the rest parameters. Thus, LSH can be applied for further speedup. 
\item It is a machine learning free (ML-free) method, which means it can be used in plug-and-play style, without requiring additional tuning processes or alignment information, once the original model is done training. 
\end{enumerate}

\begin{table}
\centering
\begin{tabular}{|c|c|c|c|}
\hline
 & \makecell{Speedup\\Train} & \makecell{Speedup \\Decode}  & \makecell{ML\\Free}\\ 
\hline
TPU & X & X & X\\ 
\hline
\makecell{Low-p} & X & X & X\\ 
\hline
\makecell{WP} &  & X & \\ 
\hline
\makecell{KD} &  & X & \\ 
\hline
NCE & X &  & n/a\\ 
\hline
\makecell{D-softmax} & X & X & X\\ 
\hline
\makecell{H-softmax}& X &  & n/a\\ 
\hline
\makecell{WA} & X & X & \\ 
\hline
LSH &  & X & X\\ 
\hline\end{tabular}
\caption{Comparison of speedup methods.}
\label{table:relatedwork}
\end{table}

\section{GPU Computing Concept}
In this section, we describe the basic concepts of CUDA-enabled GPU computing to better motivate our decisions in re-designing the LSH algorithm. 

\subsection{Warp}	
\begin{figure}[h]
  \centering
  \includegraphics[width=8.5cm]{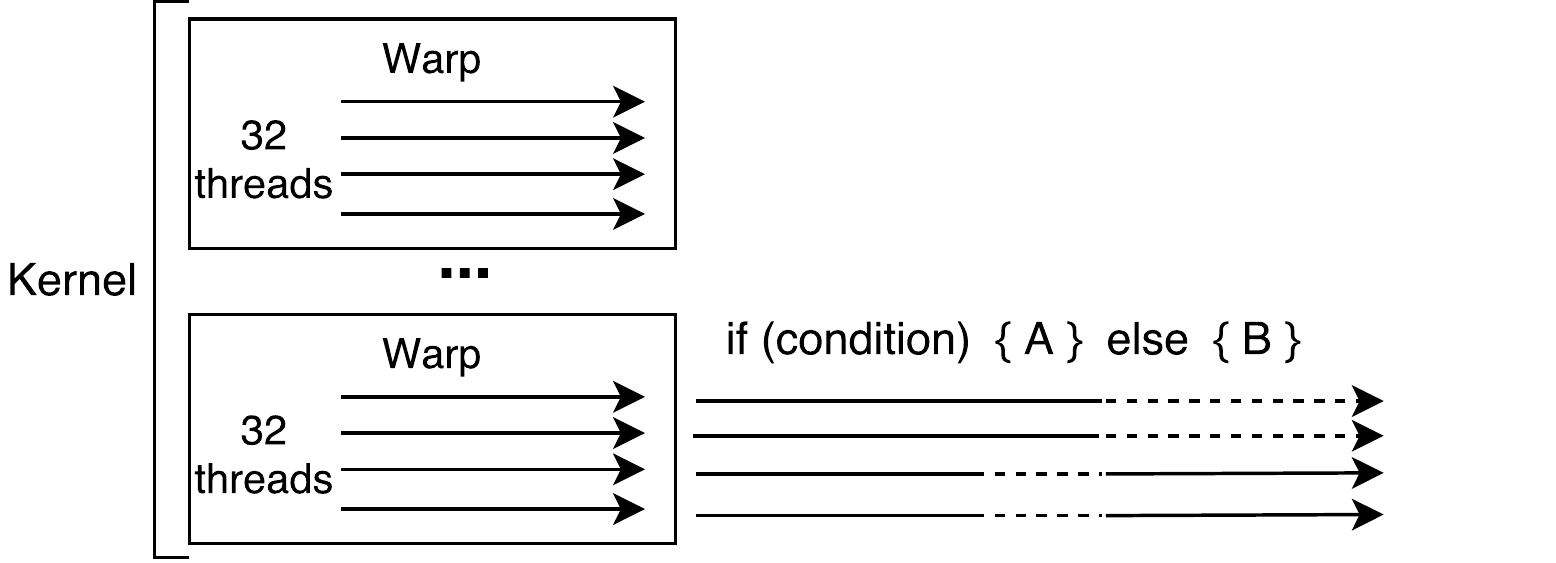}
  \caption{Illustration of kernel, warp and warp divergence. The solid line means the thread is active, and the dashed line means the thread is idle. Because of the branch, the first half of the warp will execute instruction A and be idle when the other half executes instruction B. }
  \label{fig:warp}
\end{figure}

\textit{Kernels} are functions executed on GPU. Each kernel will be executed by many GPU threads in parallel. These threads are further grouped into \textit{warps}, where each warp consists of 32 threads. All 32 threads in a wrap will execute the same instruction concurrently. However, due to the branches inside the code, some threads in a warp will be \textit{diverged} and the rest of the threads in the same wrap will have to idle in that cycle. We should make every effort to avoid warp divergence to maximize GPU usage. Figure~\ref{fig:warp} shows an example of \textit{warp divergence}. 


\subsection{Memory Hierarchy}
The bandwidth of GPU global memory is 208 GB/s for Tesla K20 GPU, and it takes 400-800 cycles for global memory access. Another faster but limited memory is \textit{shared memory}, whose bandwidth is more than 2 TB/s and only takes 3-4 cycles for each access. 

The way to access the global memory also strongly affects the speed. \textit{Coalesced access}, where  threads in the same wrap will access consecutive address, can take the full bandwidth, i.e. around 200 GB/s. Whereas the bandwidth of \textit{random access} can be as low as 20 GB/s. 

In practice, we will load data from global memory in a coalesced way to shared memory, then manipulate the data on shared memory, and finally write back to global memory in a coalesced way again.

\subsection{Latency Hiding} \label{sec:latency}

The global memory access can lead to a large latency (400-800 cycles). However, the GPU scheduler has a smart strategy to hide the latency: when a warp needs to access global memory, GPU will put it into wait, and switch to another warp to execute. In order to hide the latency completely, each kernel should launch enough threads (more than 1000) in parallel. 

\section{Locality Sensitive Hashing}

\begin{figure*}[h]
  \centering
  \includegraphics[width=14cm]{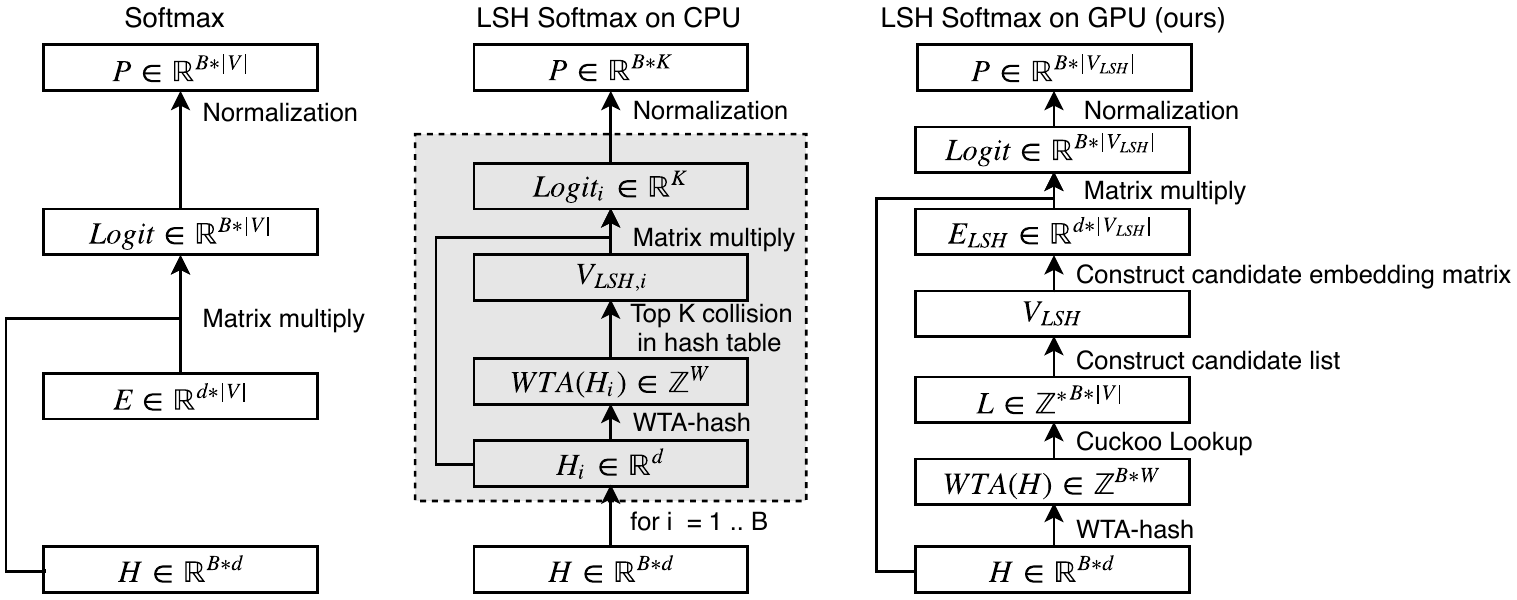}
  \caption{Comparison of the pipeline of full vocabulary softmax, LSH softmax on CPU proposed in \citeauthor{vijayanarasimhan2014deep}, and our LSH softmax on GPU. Every step of full vocabulary softmax and our LSH softmax on GPU is executed in batch mode, whereas the steps inside the grey box of LSH softmax on CPU are executed separately for each hidden vector in the beam.}
  \label{fig:pipeline}
\end{figure*}

At each step during beam search with beam size $B$, given the hidden state $H \in \mathbb{R}^{B*d}$ from the top RNN layer, the probability distribution $P \in \mathbb{R}^{B*|V|}$ over $V$ will be calculated by \textit{softmax}:
\begin{flalign}
P[i,j] &= p(y = j|H_i) = \frac{e^{Logit[i,j]}}{\sum_{k=1}^{V} e^{Logit[i,k]}} \label{eqn:softmax}\\
Logit &= H * E  \label{eqn:logit}
\end{flalign}
where $H_i$ is the ith row of $H$ and $E \in \mathbb{R}^{d*|V|}$ is the word embedding matrix. The computational intensive part is the matrix product in \ref{eqn:logit}, whose complexity is $\mathcal{O}{(dB|V|)}$. 

For each beam entry, we are only interested in the top $B$ words according to the probability/logit. We can reduce the complexity down to $\mathcal{O}{(dBV')} (|V'| \ll |V|)$ if we can inexpensively construct a much smaller vocabulary set $V'$ that also contains the top $B$ words. 

$p(y=j|H_i)$ is proportional to the dot-product between $H_i$ and $E_j$. Thus, finding the top $B$ words with highest probability is equivalent to finding the nearest neighbors of $H_i$ from all embedding vectors $E_k, \forall k = 1 ... |V|$ under the dot-product distance measure. LSH \cite{gionis1999similarity} is an efficient tool for the nearest neighbor problem. LSH will construct a small candidate vocabulary set $V_{LSH}$ which will contains the top $B$ words with a high expectation. 

\subsection{LSH on CPU}

\citeauthor{vijayanarasimhan2014deep} successfully utilize winner-take-all (WTA) LSH to speed up the softmax calculation on CPU:

\begin{enumerate}
\item Hash every word embedding $E_j \in \mathbb{R}^{d}$ into hash code $WTA(E_j) \in \mathbb{Z}^{W}$ where $W$ is the dimension of the hash space. Organize these hash codes in hash tables. 
\item For each hidden vector $H_i \in \mathbb{R}^{d}$, apply the same hash function to get $WTA(H_i)$.
\item Given $WTA(H_i)$, select the top $K$ collisions in the hash table, to construct the candidate vocabulary set $V_{LSH, i}$.
\item Calculate the dot-product $H_i * E_k, \forall k \in V_{LSH,i}$.
\item Repeat step 2-4 for each entry in the batch. 
\end{enumerate}

The tasks \citeauthor{vijayanarasimhan2014deep} that sped up are Image Classification, Skipgram Word2Vec, and Video Identification, which all involve only one step of the softmax calculation. When conducting inference in batch mode, the test entries inside a batch can be very different, thus the candidate list $V_{LSH,i}$ will differ a lot. Therefore, step 2-4 must be executed independently for each entry in the batch. 

This will lead to less speedup when batch size is large: \citeauthor{vijayanarasimhan2014deep} reports that speedup decreases from 6.9x to 1.6x when batch size increase from 8 to 64. This is problematic as beam size could be more than 50 for certain task. 

\subsection{LSH on GPU}

\begin{table}
\centering
\begin{tabular}{|c|c|c|}
\hline
(m,k,n): $A^{m,k} * B^{k,n}$ & Time (ms) & Gflop/s\\ 
\hline 
(12,1000,50000) & 5.12 & 234.58\\ 
\hline
(1,1000,50000) & 1.63 & 61.27\\ 
\hline\end{tabular}
\caption{The time consumption and floating point operations per second(Gflop/s) of matrix multiplication on GPU at different scales. } 
\label{table:gpumatrix}
\end{table}

A similar reduction in speedup will also happen on GPU, especially because GPUs prefer large-scale calculation to hide latency, as described in Section~\ref{sec:latency}. Table~\ref{table:gpumatrix} shows a comparison of matrix multiplications at different scales. Even though calculation is reduced to 1/12th, the time spent is only shrunk to one third. 

On GPU, another drawback of processing each beam one by one, is that we must first construct embedding matrix $E_{LSH,i}$ that occupies contiguous space in global memory to do matrix multiplication. This expensive data movement can further downgrade the speedups. 

To solve this problem, we propose to share the candidate vocabulary set $V_{LSH} = \cup_{i=1}^{B} V_{LSH,i}$ across different entries in the batch. Only one embedding matrix $E_{LSH}$ will be constructed and the following matrix multiplication $H * E_{LSH}$ will be calculated in batch mode in a single kernel launch. This idea is motivated by the intuition that during beam search, at each step, different $V_{LSH,i}$ will have a big portion of words in common, thus $|V_{LSH}| < \sum_{i=1}^{B}|V_{LSH,i}|$. Although the amount computation in matrix multiplication will increase ($dB|V_{LSH}|  > \sum_{i=1}^{B}d|V_{LSH,i}|$), it makes every step of our LSH algorithm on GPU executed in batch mode, saving a lot of time in practice.  

Another issue that beam search poses is that the error will accumulate fast as the sentence is built.  Therefore, missing a correct word at a certain step will lead to a catastrophe. In practice, we find that even the combined candidate list $V_{LSH}$ can miss out some important words. We solve this problem by further merging the top $T$ frequent words into the candidate list: 
\begin{flalign}
V_{LSH} &= V_{T} \cup (\cup_{i=1}^{B} V_{LSH, i})
\end{flalign}

Figure~\ref{fig:pipeline} illustrates the detailed pipeline of our LSH algorithm on GPU. Table~\ref{table:lshbreakdown} shows the time breakdown of each step in LSH softmax. Although every step is running nicely in batch mode, we still need to carefully design each step as a naive implementation leads to large overhead on GPU. The naive implementation of \textit{cuckoo lookup} and \textit{construct candidate list} experience 3.4x and 1.7x slowdown respectively, compared to our optimized version. 

\begin{table*}
\centering
\begin{tabular}{|l|r|r|r|r|r|}
\hline
 & Full Vocab (ms) & Percent & LSH (ms) & Percent & \makecell{Naive slowdown}\\ 
\hline
Softmax & 120.97 & 100.0 \% & 44.09 & 100.0 \% & \\ 
\hline
-- LSH overhead & - &  & 16.53 & 37.5 \% & \\ 
-- -- WTA-hash & - &  & 3.72 & 8.4 \% & \\ 
-- -- Cuckoo lookup & - &  & 7.29 & 16.5 \% & 3.4x \\ 
-- -- Construct candidate list & - &  & 2.51 & 5.7 \% & 1.7x \\ 
-- -- Construct $E_{LSH}$ & - &  & 3.01 & 6.8 \% & \\ 
\hline
-- Matrix multiply & 108.16 & 89.4 \% & 22.43 & 50.9 \% & \\ 
\hline
-- Normalization  & 12.33 & 10.2 \% & 2.74 & 6.2 \% & \\ 
\hline
Runtime vocab size & 40,000 &  & 6,177 &  & \\ 
\hline\end{tabular}
\caption{The runtime vocabulary size and  time breakdown of each step of full vocabulary decoding and our LSH decoding on translating a French sentence to an English sentence with beam size 12. The last column means the slowdown if the corresponding optimized step is replaced by a naive implementation. } 
\label{table:lshbreakdown}
\end{table*}

\subsubsection{Winner-Take-All hashing} \label{sec:wta}
Following \citeauthor{vijayanarasimhan2014deep}, we use the winner-take-all (WTA) \cite{yagnik2011power} hashing function, which can be formally defined in the following equations: 
\begin{flalign}
&WTA(H \in \mathbb{R}^d) = [I_1;...;I_p;...;I_P] \\
&I_p = \operatorname{arg\,max}_{k=1}^{K} Permute_p(H)[k]
\end{flalign}

WTA will convert a $d$-dimension real value hidden vector into a $P$-dimension int value \textit{hash code}. Each $I_p$ is the index of the maximum value of the first $K$ elements of the permuted $H$. Thus, the hash code is actually an ordinal embedding of the original hidden vector, and we use the ordinal similarity as a proxy for the dot-product similarity. We can further group these hash codes into bands, and convert into a $W$-dimension \textit{band code}:  
\begin{flalign}
&WTA_{band}(H) = [B_1;...;B_w;...;B_W] \label{eqn:band} \\
&B_w = [I_{(w-1)*u + 1};...;I_{(w-1)*u + i};...;I_{w*u}]  \\
&u = P / W 
\end{flalign}
where each band code $B_w$ is the concatenation of $u$ hash codes. Table~\ref{table:wtaexample} shows an example of WTA hash. We can represent $B_w$ in $u*\log_2(K)$ bits, and we make sure $u*\log_2(K) < 31$ so that we can store each band code using an int32 on GPU. The hyper parameters for WTA hash are $\{K,u,W\}$.

\begin{table}[h]
\centering
\begin{tabular}{|c|c c c c|}
\hline
$H$ & 0.32 & 0.48 & -0.57 & 0.63\\ 
\hline
$Permute_0$ & 1 & 2 & 4 & 3\\ 
$Permute_1$ & 1 & 3 & 2 & 4\\ 
$Permute_2$ & 3 & 2 & 4 & 1\\ 
$Permute_3$ & 4 & 1 & 2 & 3\\ 
\hline
$I_1$ & 1 $\rightarrow$ $(00)_2$ &  &  & \\ 
$I_2$ & 2 $\rightarrow$ $(01)_2$& &  & \\ 
$I_3$ & 2 $\rightarrow$ $(01)_2$& &  & \\ 
$I_4$ & 1 $\rightarrow$ $(00)_2$& &  & \\ 
\hline
$B_1$ & $(0001)_2$ &  &  & \\ 
$B_2$ & $(0100)_2$ &  &  & \\ 
\hline\end{tabular}
\caption{The running example of WTA hash with $W=2$, $u=2$ and $K=2$.} 
\label{table:wtaexample}
\end{table}

\subsubsection{Cuckoo lookup}

\begin{figure}[h]
  \centering
  \includegraphics[width=7cm]{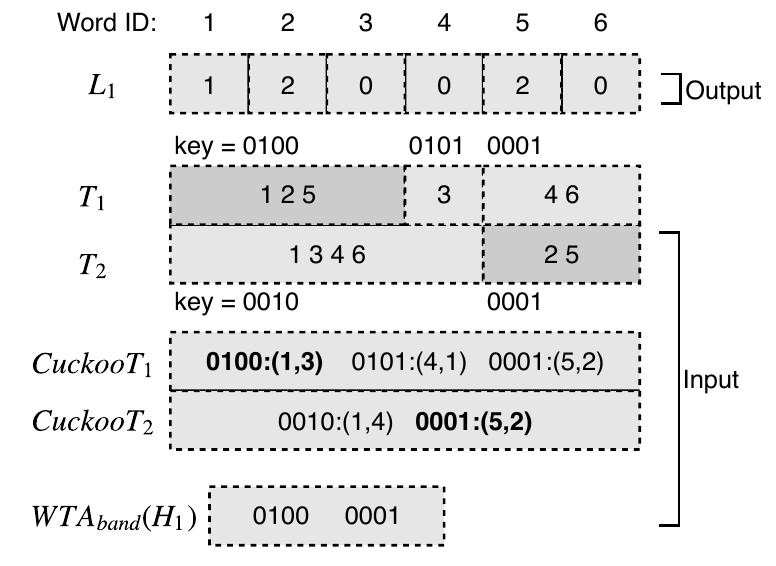}
  \caption{Example of cuckoo lookup. The beam size is 1, $W=2$ and $|V| = 6$.}
  \label{fig:cuckoo}
\end{figure}

Given $WTA_{band}(H) \in \mathbb{Z}^{B*W}$, this step will calculate the hit matrix $L \in \mathbb{Z*}^{B*|V|}$, where
\begin{multline}
L[i,j] = \\ \sum_{w = 1}^{W} I(WTA_{band}(H_i)[w] = WTA_{band}(E_j)[w])
\end{multline}
$L[i,j]$ counts how many band codes are the same between $WTA_{band}(H_i)$ and $WTA_{band}(E_j)$, which estimates the dot-product similarity between $H_i$ and $E_j$. 

First, we hash all word embeddings $WTA_{band}(E_j)$. For each band, we will build a hash table:\\
$T_w = \{ band\_code : [word\_id_1, ... word\_id_n]\}$
where the key is the band code and the value is a list containing all the words whose band code is equal to the key. We re-organize $T_w$ into an flat array on GPU: word ids with same band code are stored in continuous span, and we build a cuckoo hash table for each $T_w$ to store the starting position and corresponding length of each span:\\
$CuckooT_{w} = \{band\_code : (start,length)\}$

Second, we launch a total of $B*W$ GPU threads to calculate $L$, and each thread follows Algorithm~\ref{alg:cuckoo}. To look up certain key in cuckoo hash table, it hashes and compares the key at most twice. Thus the warp(a group of 32 threads) won't diverge at line 2. 

\begin{figure}[h]
  \centering
  \includegraphics[width=6cm]{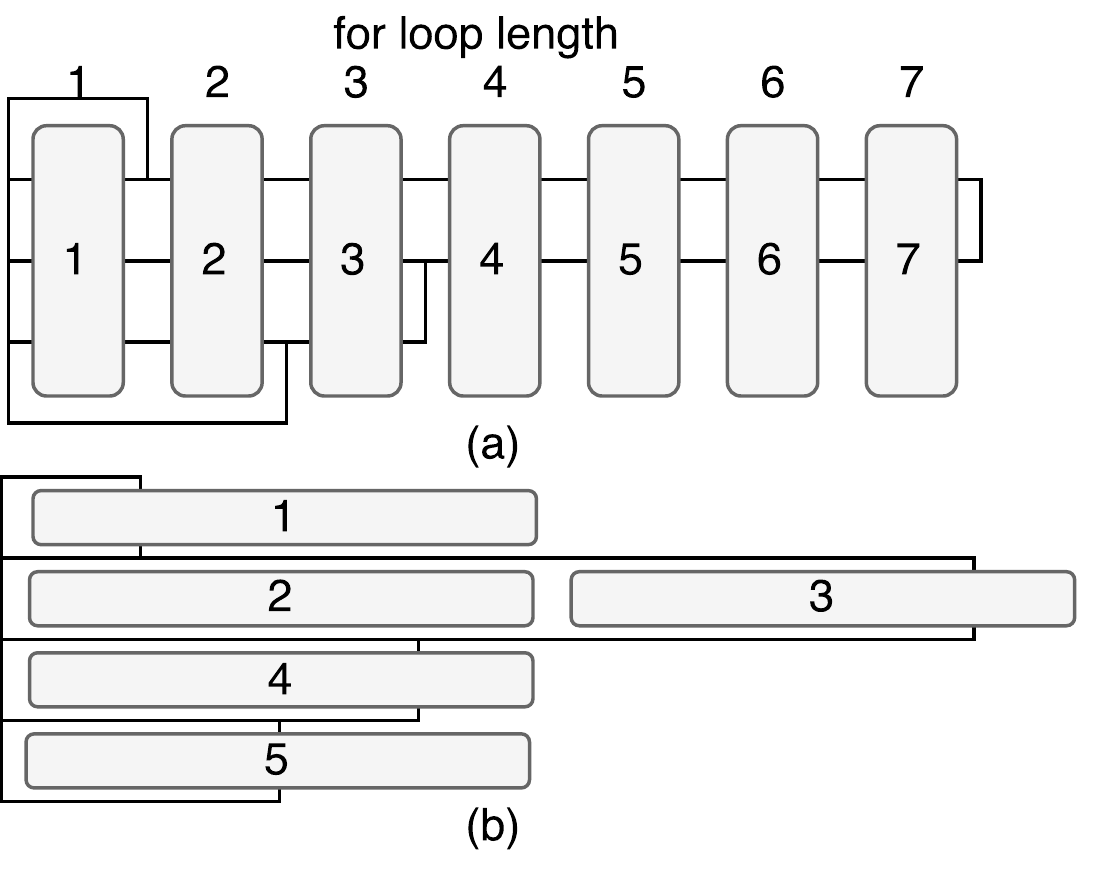}
  \caption{Illustration of naive implementation and optimized implementation of line 3-6 in Algorithm~\ref{alg:cuckoo}. We assume each warp contains 4 threads, and their for-loop lengths are 1, 7, 3, and 2. The round grey rectangle represents one step of a warp. (a) The naive implementation, which takes the warp 7 steps to finish. (b) The optimized implementation, which takes only 5 steps. }
  \label{fig:cuckoowarp}
\end{figure}

However, at line 3, because different threads will have different $length$ values, the execution time of the warp will depend on the largest length. There will be a serious \textit{warp divergence} at line 3-6. To solve this problem, we re-arrange the execution order so that the 32 threads will first process the for-loop of thread$_0$ together, then the for-loop of thread$_1$ together, until that of thread$_{31}$. Such re-arrangement will speed up this step by 3.4x. Figure~\ref{fig:cuckoowarp} illustrates the two different thread arrangements. 
\begin{algorithm}
\caption{Cuckoo lookup}
\begin{algorithmic}[1]
\item[\textbf{Inputs:}] $T$, $CuckooT$, $WTA_{band}(H)$
\item[]\ \ \ \ \ \ \ beam index $i$, band index $w$
\item[\textbf{Output:}] $L$
\STATE code = $WTA_{band}(H_i)[w]$
\STATE start, length = $CuckooT_w[code]$
\FOR{pos = start to start + length}
\STATE word\_id = $T_w[pos]$
\STATE L[i,word\_id] += 1
\ENDFOR
\end{algorithmic}
\label{alg:cuckoo}
\end{algorithm}

\subsubsection{Construct candidate list}

Given the hit matrix $L \in \mathbb{Z*}^{B*|V|}$ and a threshold $t$, this step selects the final candidate vocabulary set $V_{LSH}$, where:
\begin{equation}
j \in V_{LSH} \iff \exists i, s.t. L[i,j] >= t
\end{equation}

\begin{figure}[h]
  \centering
  \includegraphics[width=8cm]{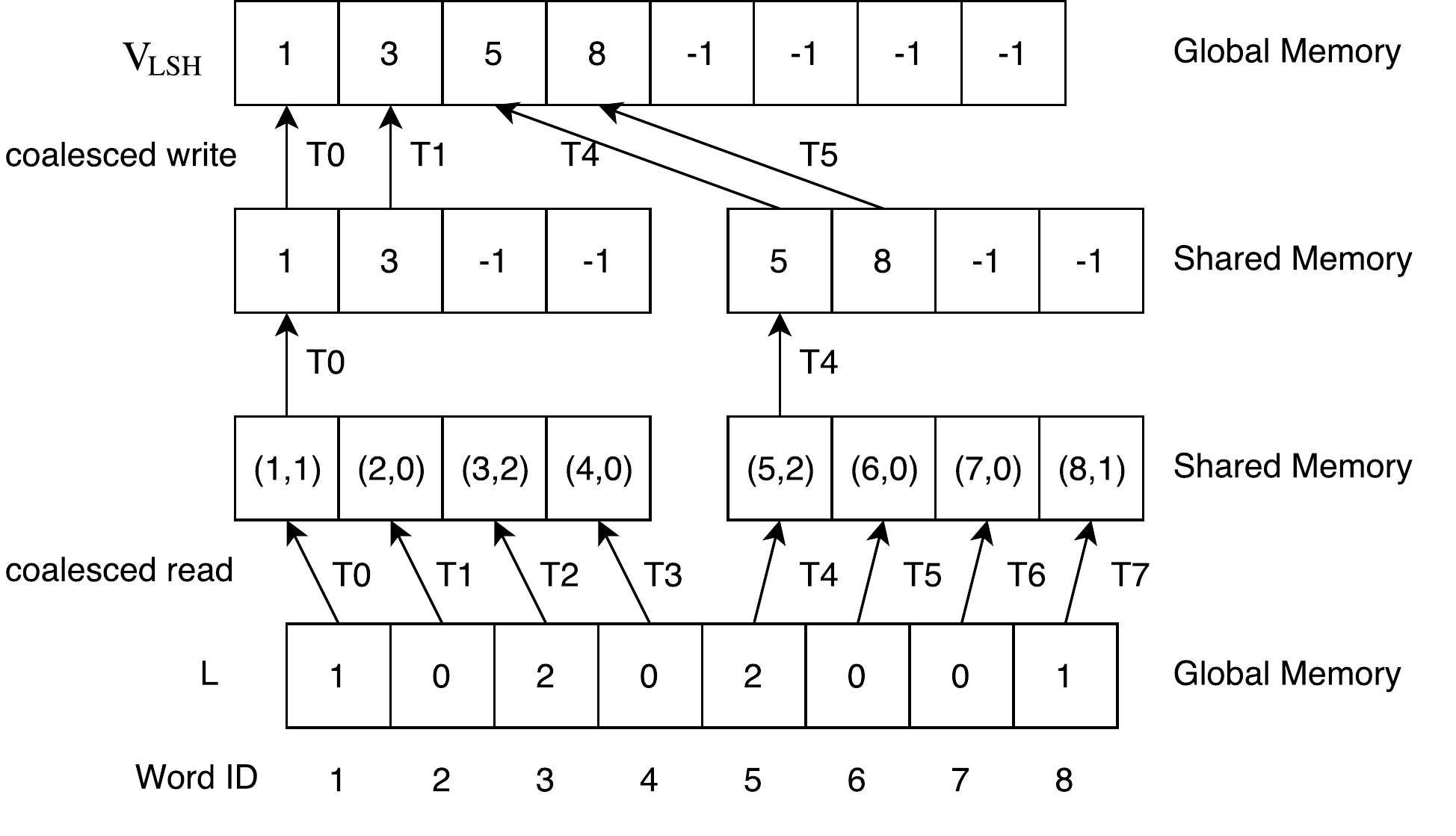}
  \caption{Illustration of optimized stream compaction algorithm. We assume each warp contains 4 threads here. 2 warps will first load $L$ into shared memory in a coalesced read. Then only the first thread of each warp will scan the 4 values and filter out the valid word ID. Then each warp will write the valid word ID back in a coalesced write. The start position in $V_{LSH}$ for each warp is maintained in global memory, omitted here.}
  \label{fig:vlsh}
\end{figure}

We use threshold to avoid the inefficient sorting on GPU. $L$ is a sparse matrix after filtering with $t$, whereas $V_{LSH}$ should be a dense array. This is the canonical \textit{Stream Compaction} problem, and one can simply use \textit{copy\_if} function provided in \textit{thrust} library. To further improve the efficiency, we re-design the algorithm by taking advantage of \textit{shared memory} and \textit{coalesced access}. The new algorithm is illustrated in Figure~\ref{fig:vlsh}. 

Hyper-parameters that define a WTA LSH beam search are $\{K,u,W;B,T,t\}$, where $B$ is beam size, $T$ is the number of top frequent words to merge and $t$ is the threshold to select $V_{LSH}$.

\section{Experiment} 

\begin{table}
\centering
\begin{tabular}{|c|c|c|c|c|}
\hline
 & J2E & E2J & F2E & U2E\\ 
\hline
$|V_{source}|$ & 80K & 88K & 200K & 50K\\ 
\hline
$|V_{target}|$  & 50K & 66K & 40K & 25K\\ 
\hline
\#Tokens & 70.4M & 70.4M & 652M & 3.3M\\ 
\hline
Attention & Yes & Yes & No & Yes\\ 
\hline\end{tabular}
\caption{Training configurations of different language pairs. The attention model is based on \citeauthor{luong2015effective}. Data sources: ASPEC Japanese-English Corpus \cite{NAKAZAWA16.621}, French-English Corpus from WMT2014 \cite{wmt2014}, and Uzbek-English Corpus \cite{bolt2016}.}
\label{table:dataset}
\end{table}

We conduct our experiment on 4 machine translation models: Japanese to English (J2E), English to Japanese (E2J), French to English (F2E) and Uzbek to English (U2E). The statistics and training parameters are shown in Table~\ref{table:dataset}.

\begin{table}
\centering
\begin{tabular}{|c|c|c|c|c|}
\hline
 & J2E & E2J & F2E & U2E\\ 
\hline
Softmax Speedup & 3.95 & 2.46 & 3.22 & 2.04\\ 
\hline
Overall Speedup & 2.12 & 2 & 2.05 & 1.78\\ 
\hline
BLEU Loss & 0.43 & 0.07 & 0.31 & -0.28\\ 
\hline
\end{tabular}
\caption{The speedup of softmax module and overall pipeline of LSH decoding over full softmax decoding.} 
\label{table:speedups}
\end{table}

\textbf{Overall speedup} As Table~\ref{table:speedups} shows, our LSH decoding  achieves up to 4x speedup on the softmax, and a consistent 2x overall speedup for J2E, E2J and F2E with tiny BLEU score loss. For U2E, the speedup without BLEU loss is 1.78x, due to the small original target vocabulary size (25,000).  

\begin{figure}[ht]
  \centering
  \includegraphics[width=8cm]{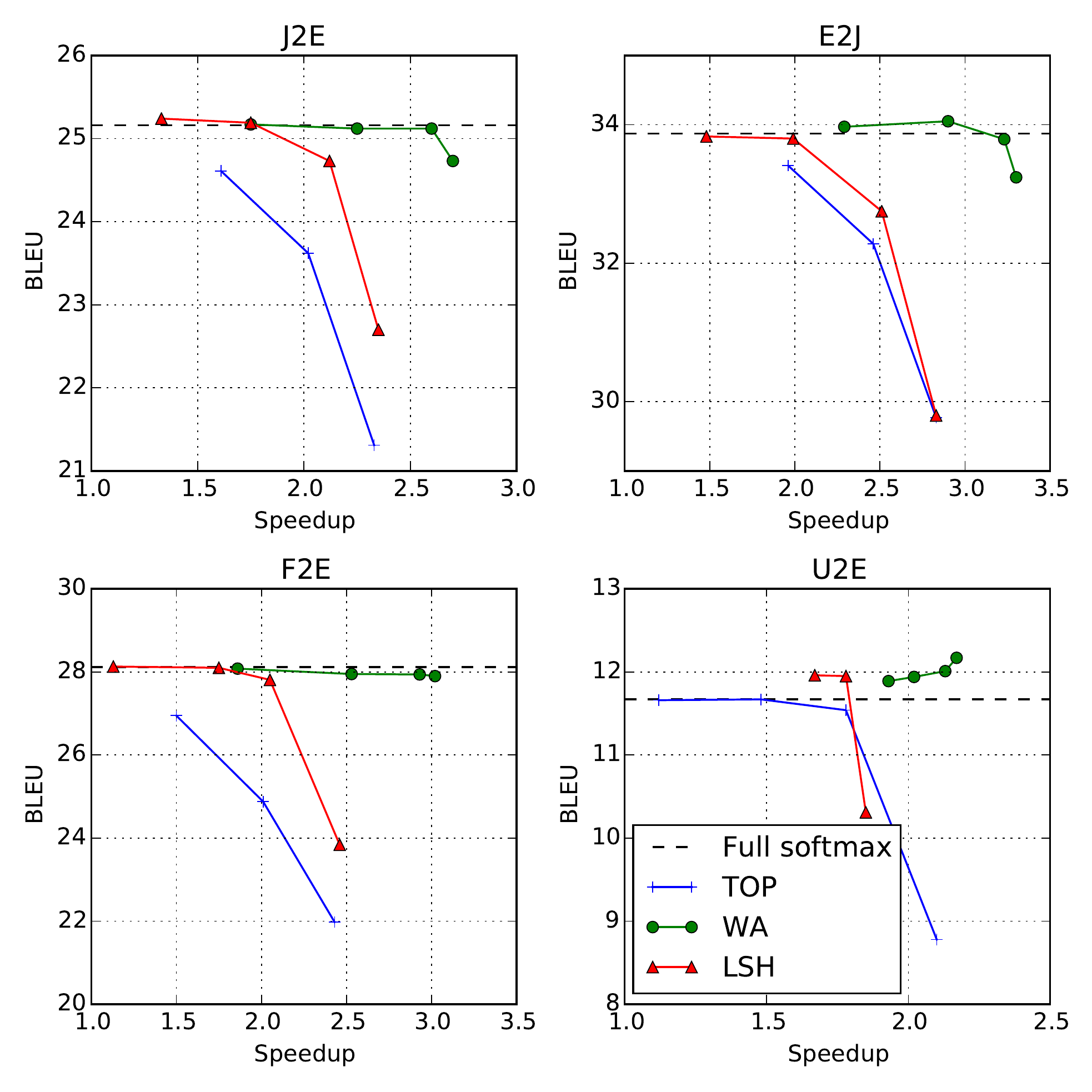}
  \caption{BLEU/speedup curve of 3 decoding methods on 4 translation directions. For decoding with top frequent words (TOP), we vary the number of top frequent words to get different BLEU/speedup. For decoding with word alignment (WA), we vary the number of aligned target words of each source word. For LSH decoding, we vary both the number of top frequent words to merge ($T$) and the threshold ($t$). }
  \label{fig:speedup}
\end{figure}

We compare our algorithm with two other decoding acceleration methods: Decoding using only the top frequent words (TOP) and decoding with word alignments (WA) \cite{shi2017speeding}. We conduct a grid search of the LSH hyper parameters $\{K,u,W\}$ and find that $\{8,3,500\}$ and $\{16,3,500\}$ generally deliver good performance/speedup trade-off. We vary other two hyper parameters $\{T, t\}$ to get different speedup and BLEU score. Figure~\ref{fig:speedup} shows the BLEU/speedup curve of the three decoding methods on 4 translation directions. 

Our LSH decoding always obtain a higher BLEU with a large margin than TOP decoding at the same speedup level. Table~\ref{table:lshbreakdown} shows that the optimized LSH overhead can take up to 37.5\% of the total time to calculate softmax. Thus, to achieve the same speedup with TOP decoding, the runtime vocabulary size of LSH can not exceed half of that of TOP decoding, which demonstrates that our LSH algorithm indeed selects a smaller yet more accurate vocabulary set. 

The WA decoding achieves higher speedup with the same BLEU. However, this approach can only work in the context where a sensible alignment information can be provided.  

\begin{figure}
  \centering
  \includegraphics[width=8cm]{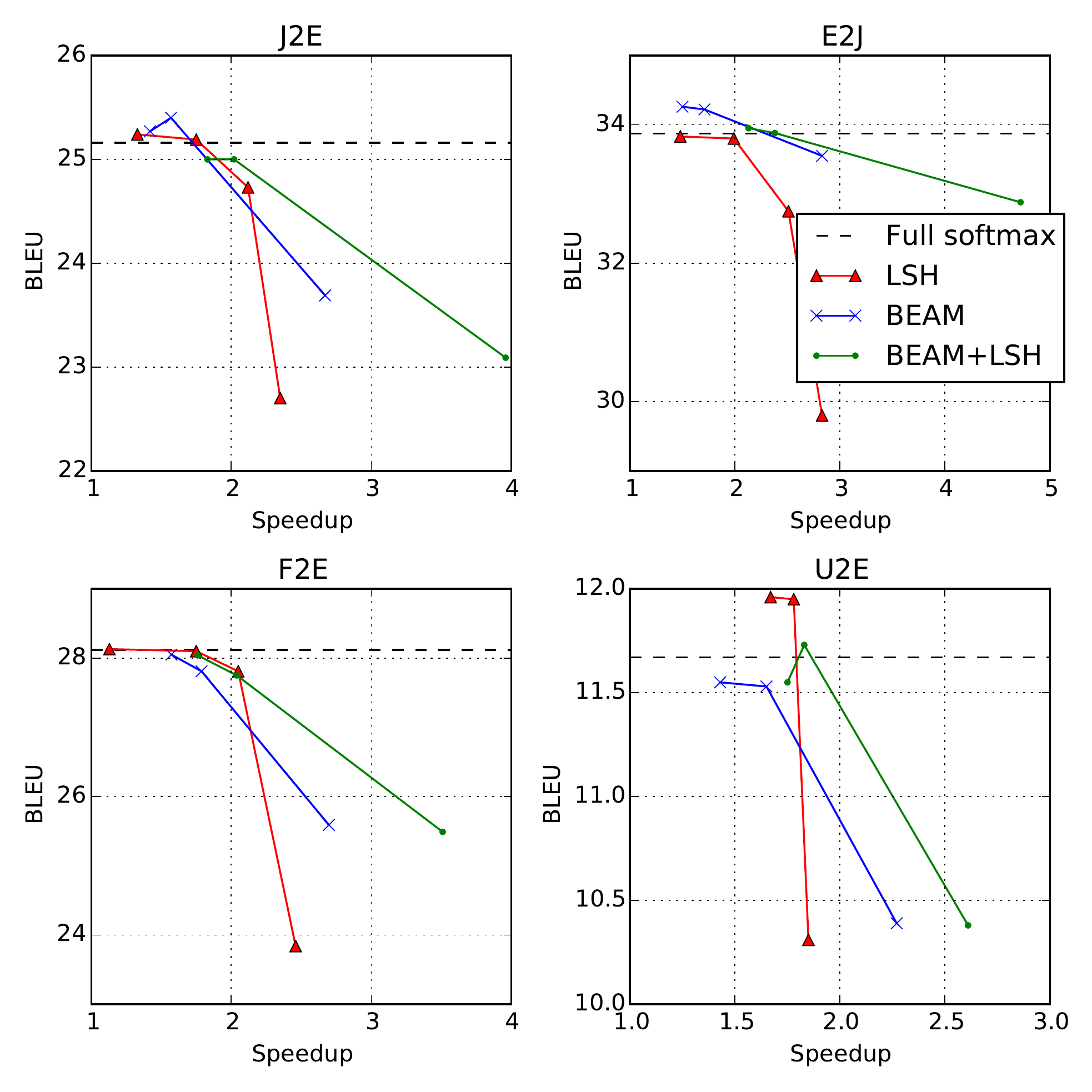}
  \caption{BLEU/speedup curve of LSH decoding (LSH), full softmax decoding with smaller beam size (BEAM) and LSH decoding with smaller beam size (BEAM+LSH). All the speedups are calculated over the speed of full softmax decoding with beam size 12. The three smaller beam sizes are 6, 3 and 1. }
  \label{fig:speedup_beam}
\end{figure}

\textbf{Effects of beam size} 
Another easy way to speed up decoding is to just reduce the beam size, which is also orthogonal to our LSH decoding. Figure~\ref{fig:speedup_beam} demonstrates that LSH decoding with reduced beam size can achieve even better speed/performance trade-off. 

\begin{table}
\centering
\begin{tabular}{|c|c|c|}
\hline
Beam size & Speedup  & BLEU loss\\ 
\hline
12 & 2.06 & 0.31\\ 
\hline
24 & 2.22 & 0.35\\ 
\hline
36 & 2.23 & 0.28\\ 
\hline
48 & 2.21 & 0.31\\ 
\hline\end{tabular}
\caption{The speedup and BLEU loss of LSH decoding over full softmax decoding at different beam sizes on F2E.} 
\label{table:beameffect}
\end{table}

On the other hand, the speed and performance at larger beam size is also important because certain application, like poem generation, requires the beam size larger than 50 to work in practice. Table~\ref{table:beameffect} shows the speedup and BLEU loss of our LSH decoding over the full vocabulary decoding at larger beam size (batch size). Unlike the algorithm proposed by \citeauthor{vijayanarasimhan2014deep}, our algorithm will maintain or even obtain a higher speedup with the same level of BLEU when beam size (batch size) increases, which can be further explained by that large batch size will saturate the GPU and fully exploit its parallel power. 

\begin{figure}[b]
  \centering
  \includegraphics[width=5cm]{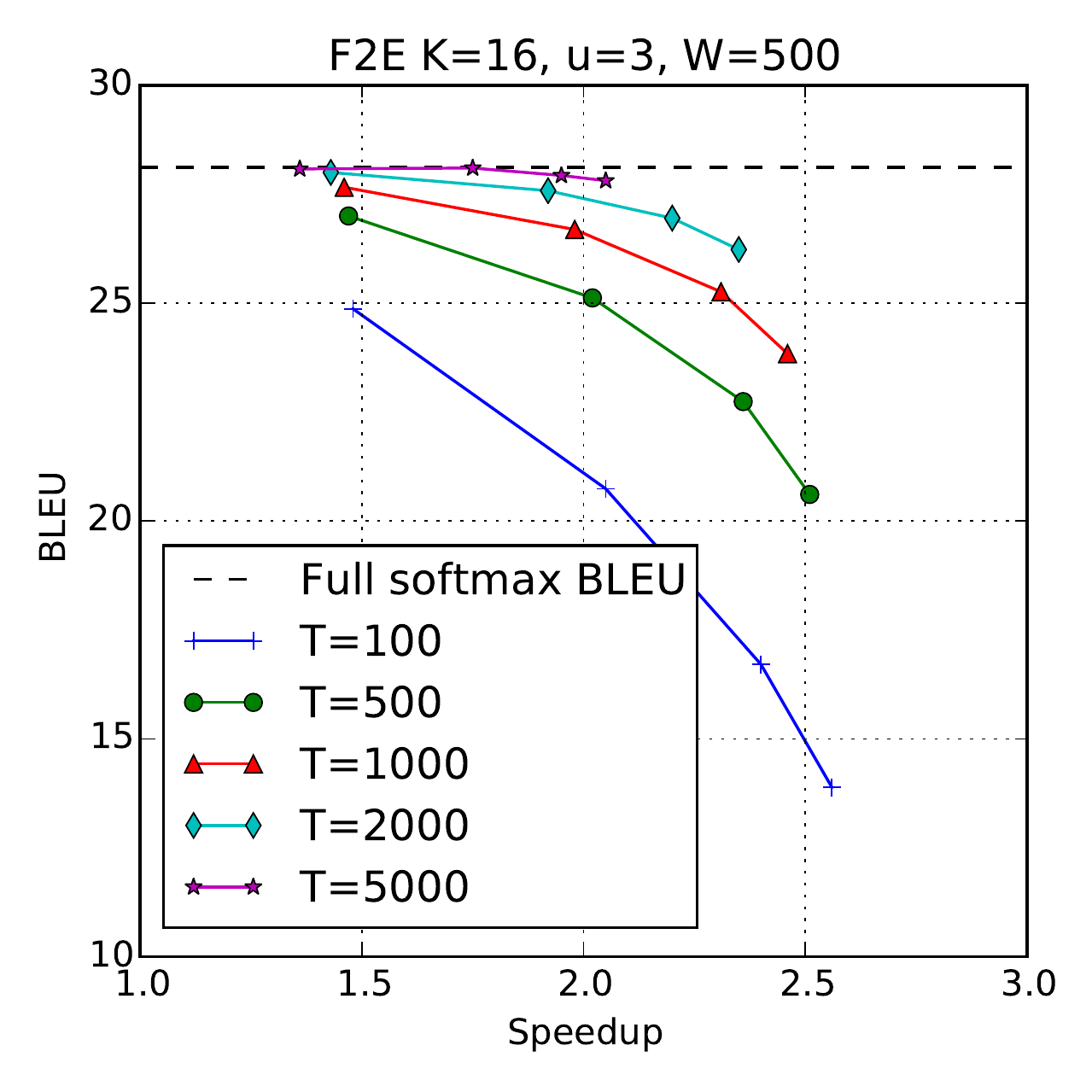}
  \caption{The BLEU/speedup curve for different $T$ on French to English translation model.}
  \label{fig:speedup_T}
\end{figure}

\textbf{Effects of T} The merge of top $T$ frequent words into the candidate word list $V_{LSH}$ is necessary to obtain good performance. Figure~\ref{fig:speedup_T} shows the BLEU/speedup curve for different $T$ on the French-to-English translation task. Having $T$ too small will result in low BLEU whereas having $T$ too large will limit the highest speedup the method can achieve.


\section{Conclusion}
We re-design the LSH algorithm for beam search on GPU. The candidate vocabulary set $V_{LSH}$ is shared across the beams to execute every step in batch mode. Several key functions are optimized by using a cuckoo hash table, taking advantage of shared memory, and avoiding warp divergence. Top frequent words are merged into $V_{LSH}$ to further improve the performance. Our LSH algorithm is a machine-learning-free acceleration method that achieves 2x speedup on 4 machine translation tasks, and delivers better BLEU/speedup trade-off than TOP decoding.

\bibliography{emnlp2017}
\bibliographystyle{aaai}

\end{document}